\documentclass[pdflatex,sn-mathphys-num]{sn-jnl}

\usepackage{graphicx}%
\usepackage{multirow}%
\usepackage{amsmath,amssymb,amsfonts}%
\usepackage{amsthm}%
\usepackage{mathrsfs}%
\usepackage[title]{appendix}%
\usepackage{xcolor}%
\usepackage{textcomp}%
\usepackage{manyfoot}%
\usepackage{booktabs}%
\usepackage{algorithm}%
\usepackage{algorithmicx}%
\usepackage{algpseudocode}%
\usepackage{listings}%

\theoremstyle{thmstyleone}%
\theoremstyle{thmstyletwo}%

\theoremstyle{thmstylethree}%

\begin{document}

\title[Article Title]{Training Kindai OCR with parallel textline images and self-attention feature distance-based loss}


\author*[1]{\fnm{Anh} \sur{Le}}

\author[2]{\fnm{Asanobu} \sur{Kitamoto}}




\abstract{Kindai documents, written in modern Japanese from the late 19th to early 20th century, hold significant historical value for researchers studying societal structures, daily life, and environmental conditions of that period. However, transcribing these documents remains a labor-intensive and time-consuming task, resulting in limited annotated data for training optical character recognition (OCR) systems. This research addresses this challenge of data scarcity by leveraging parallel textline images—pairs of original Kindai text and their counterparts in contemporary Japanese fonts—to augment training datasets. We introduce a distance-based objective function that minimizes the gap between self-attention features of the parallel image pairs. Specifically, we explore Euclidean distance and Maximum Mean Discrepancy (MMD) as domain adaptation metrics. Experimental results demonstrate that our method reduces the character error rate (CER) by 2.23\% and 3.94\% over a Transformer-based OCR baseline when using Euclidean distance and MMD, respectively. Furthermore, our approach improves the discriminative quality of self-attention representations, leading to more effective OCR performance for historical documents.}




\maketitle

\section{Introduction}\label{sec1}

Since historical documents are an invaluable resource for historians in exploring social aspects, daily life, politics, and even climate patterns of past eras, many countries have prioritized their preservation. A common approach is the creation of digital libraries that store high-resolution scans of these documents and make them accessible to both domestic and international researchers. Traditionally, the digitization process involved scanning printed or handwritten texts into image format, followed by manual transcription by experts. However, this method is extremely time-consuming, labor-intensive, and impractical for the vast collections housed in archives and libraries. To address these challenges, document analysis and recognition technologies have emerged as vital tools. These systems leverage machine learning and computer vision to automatically detect layout structures, recognize handwritten or printed characters, and convert scanned pages into machine-readable text. This automation significantly accelerates the digitization workflow, reduces dependence on scarce experts, and makes large-scale transcription and analysis feasible, thus greatly enhancing access to historical knowledge for scholars around the world.

In Japan, historical documents from the late 19th to early 20th century are classified as Kindai documents, referring to the "modern" period just before full modernization. These materials are critically important for scholars studying Japan’s transition into a modern state, and they encompass a wide range of content, including government records, newspapers, magazines, books, and legal documents. Due to their historical and cultural significance, digitizing and making these documents machine-readable is a priority. However, this task presents significant technical challenges.

Document recognition systems for Kindai materials typically operate in two main stages: textline detection and textline recognition. In the first stage, the system analyzes scanned pages to identify and isolate individual lines of text. Then, during recognition, each line is analyzed and converted into character codes, often in Unicode. This process is more complicated than it sounds, especially with Kindai documents, which frequently exhibit complex layouts such as multi-column formats, vertical writing, annotations in side margins. Furthermore, Kindai texts use varied typography, antiquated kanji characters, and orthographic variations, including obsolete kana usage and stylistic ligatures that are uncommon or absent in contemporary writing systems.

\begin{figure}[h]
\centering
\includegraphics[width=0.7\textwidth]{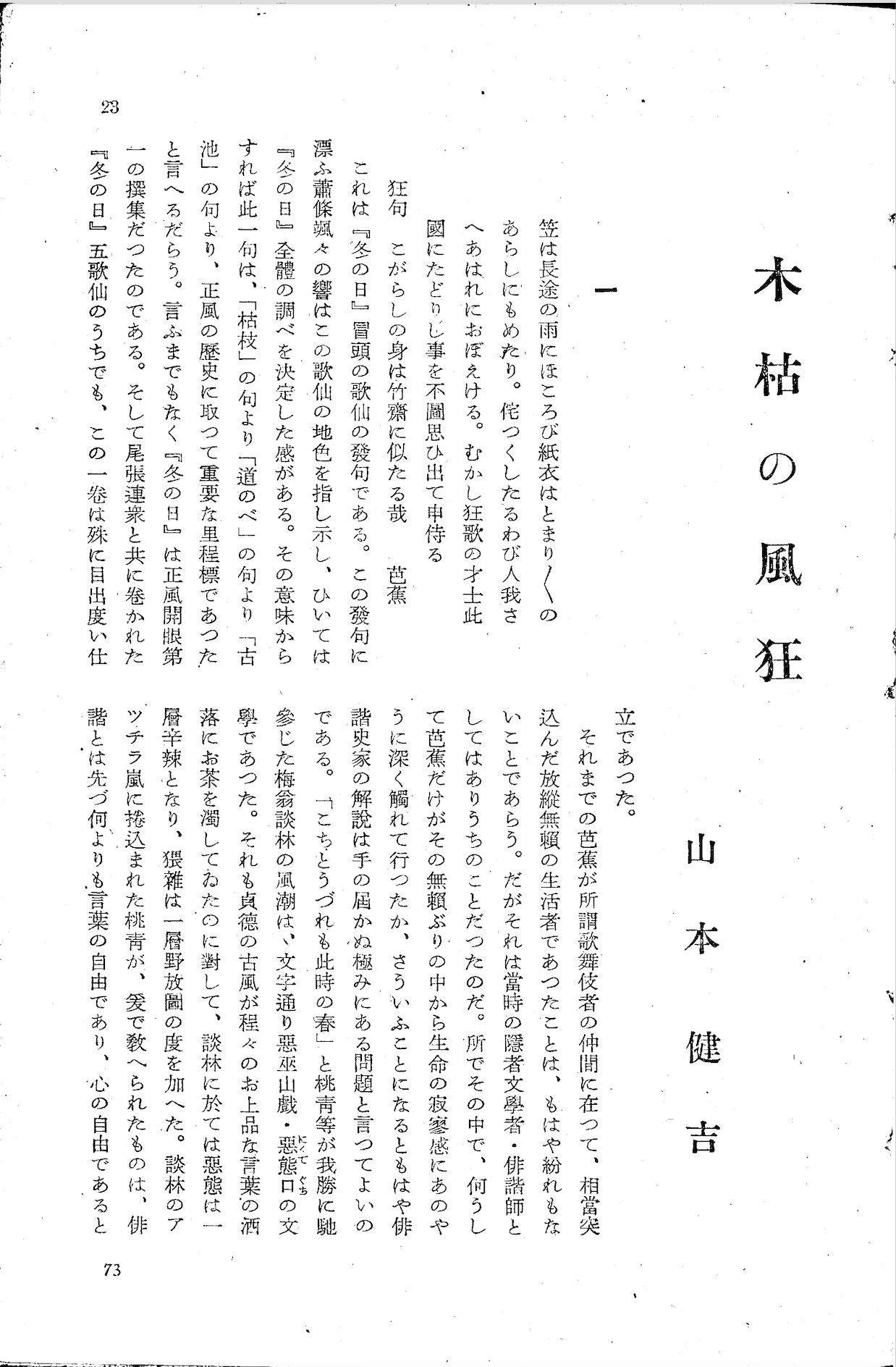}
\caption{Example of Kindai document.}\label{fig1}
\end{figure}

Another primary challenge in processing historical Japanese documents stems from the visual similarity among certain characters and the extensive vocabulary, which includes numerous kanji characters, many of which are rare or not included in modern standardized character sets. This results in a high incidence of out-of-vocabulary (OOV) characters, posing significant difficulties for conventional optical character recognition (OCR) systems. A further impediment to progress is the limited availability of large-scale, high-quality annotated datasets. For example, the publicly accessible dataset provided by the National Diet Library of Japan comprises only approximately 4,000 pages of annotated Kindai-era materials. This volume is inadequate for training robust deep learning models, which typically require tens or hundreds of thousands of labeled instances to achieve satisfactory generalization. Figure~\ref{fig1} presents an example of a Kindai document, where both the title and the main text are written in vertical script.

To address this data scarcity, one promising direction of research is to leverage synthetic data generated from modern Japanese fonts. By using font files to generate artificial textline images that resemble real Kindai textlines in style and complexity, researchers can train models in a supervised setting without requiring extensive manual annotation. The core idea is to perform domain adaptation, where the model learns useful patterns from synthetic font-based images and then applies this knowledge to recognize real Kindai document images. This approach enables the development of more robust and accurate recognition systems, even with limited real-world training data. Such cross-domain learning strategies, if further refined, could significantly enhance the usability and accessibility of Kindai documents. In turn, this would accelerate scholarly research across disciplines such as history, literature, linguistics, and political science by making vast volumes of previously unreadable texts searchable, translatable, and analyzable in digital form.

In this paper, we present a novel approach to enhancing the performance of text recognition systems for Kindai documents by addressing the critical limitation of insufficient annotated training data. Specifically, we target on improving the text recognition stage of the OCR pipeline by leveraging parallel textline images—a curated dataset composed of pairs of corresponding textline images from original Kindai documents and synthesized counterparts rendered using contemporary Japanese fonts. These parallel pairs  serve as a bridge between the scarce Kindai data and the more readily available modern Japanese font resources.

To effectively utilize these parallel images for training, we propose a distance-based objective function operating on the self-attention features extracted by the Transformer-based OCR model. The core idea is to minimize the feature-space distance between each pair of parallel textline images. By aligning the internal representations of historical and synthetic textlines, the model is encouraged to learn domain-invariant features, enabling better generalization to real Kindai text under low-resource conditions. This adaptation framework allows the model to effectively benefit from the large corpus of synthetic textlines generated from modern Japanese fonts, thereby mitigating the data scarcity issue without requiring extensive manual annotation of historical texts.

Our experiments demonstrate the effectiveness of this method in improving recognition accuracy on Kindai textline images. Incorporating both the parallel textline dataset and the proposed distance-based training objective, our model outperforms baselines that rely solely on either synthetic or limited real data. Furthermore, qualitative results show improved character-level accuracy, particularly in handling complex kanji and stylistic variations characteristic of Kindai documents.

The remainder of this paper is organized as follows: Section~\ref{sec2} reviews related work in the areas of historical Japanese document recognition and domain adaptation techniques. Section~\ref{sec3} describes the architecture of our Transformer-based OCR model, the generation process of parallel textline images, and the formulation of our distance-based objective function. Section~\ref{sec4} presents experimental setups and results. Finally, Section~\ref{sec5} concludes the paper and outlines potential directions for future work.

\section{Related Works}\label{sec2}

For training the Kindai OCR system, we utilized the Shisou dataset, comprising 1,000 annotated images. This dataset was provided by the Center for Research and Development of Higher Education at the University of Tokyo. However, it is important to note that the Shisou dataset is not publicly available due to licensing restrictions \cite{bib3}. Subsequently, the National Diet Library (NDL) released the first open dataset for Kindai documents, referred to as the NDL dataset \cite{bib1}. This dataset consists of 3,997 annotated page images, including a total of 10,421 horizontal textlines and 92,835 vertical textlines. The availability of the NDL dataset represents a significant step forward in supporting reproducible research and development in historical Japanese document recognition.

In parallel with these dataset releases, the Research Organization of Information and Systems – Data Science Center for Open Data in the Humanities (ROIS-DS CODH) has developed and released a comprehensive OCR system specifically designed for historical Kindai documents \cite{bib9}. This system includes both textline detection and textline recognition components. For the detection stage, it utilizes the CRAFT (Character Region Awareness for Text Detection) model, which is well-suited to identifying irregularly shaped and variably spaced text regions. For textline recognition, the system employs an attention-based encoder-decoder architecture \cite{bib6}, which has proven effective for handling complex script patterns and character sequences found in historical Japanese documents. The entire pipeline is trained on the Shisou dataset, which contains manually annotated images of Kindai documents.

In a related development, the National Diet Library (NDL) released its own OCR system for Kindai documents in 2022. This system extends beyond basic text recognition by incorporating a layout analysis module alongside textline recognition. The layout analysis component employs the MMDetection framework to identify and segment structural elements such as paragraphs and textlines. For the recognition stage, NDL's system utilizes a Convolutional Recurrent Neural Network (CRNN) model \cite{bib6}, a widely used architecture for sequence-based text recognition. Together, these advances represent significant progress in the automated transcription and analysis of Kindai documents, enhancing broader access and research on historical Japanese texts.

For object recognition, the Syn2Real is design to help models learn to generalize from synthetic images to real images \cite{bib11}. The dataset contains synthetic images rendered from 3D object models and real images from the same object categories. It is designed for domain adaptation with different tasks such as closed set object classification, open set object classification, and object detection. Depending on the type of data available from the target domain, domain adaptation for object recognition can be classified into three types: Supervised, Semi Supervised, Unsupervised domain adaptation. For supervised domain adaptation, a classification loss and a contrastive semantic alignment loss are proposed \cite{bib10}. The semantic alignment loss (in addition to the traditional classification loss) encourage samples from different domains belonging to the same category to map nearby in this embedding space. Maximum Mean Discrepancy (MMD) \cite{bib2} is a popular choice in domain adaptation to align the distributions of the features in the embedding space mapped from the source and the target domains. In this paper, we extend MMD to align two distributions of sequential data.

For text recognition, SynthText is a large scale dataset that contains about 800K synthetic images for text detection/recognition in the wild \cite{bib4}. These images are created by blending natural images with text rendered with random fonts, sizes, colours, and orientations. As a result, they closely resemble realistic images. The Sequence-to-Sequence Domain Adaptation Network is proposed for adapting different domains of text image recognition \cite{bib8}. A gated attention similarity unit is proposed to align the distribution of the source and target sequence data in the character-level feature space, rather than a global coarse alignment.
In addition to historical OCR datasets, prior work has used font-generated data for historical script recognition. A study on Nom documents \cite{bib12} built a digitization system with binarization, segmentation, and recognition, training on 7,601 and 32,733 categories using GLVQ and MQDF2. Without ground-truthed samples, training data was generated from 27 Chinese, Japanese, and Nom fonts, demonstrating the effectiveness of font-based data generation for scarce historical datasets.

\section{Proposed Domain Adaptation Method}\label{sec3}

\subsection{Transformer OCR}\label{subsec2}
\begin{figure}[h]
\centering
\includegraphics[width=0.5\textwidth]{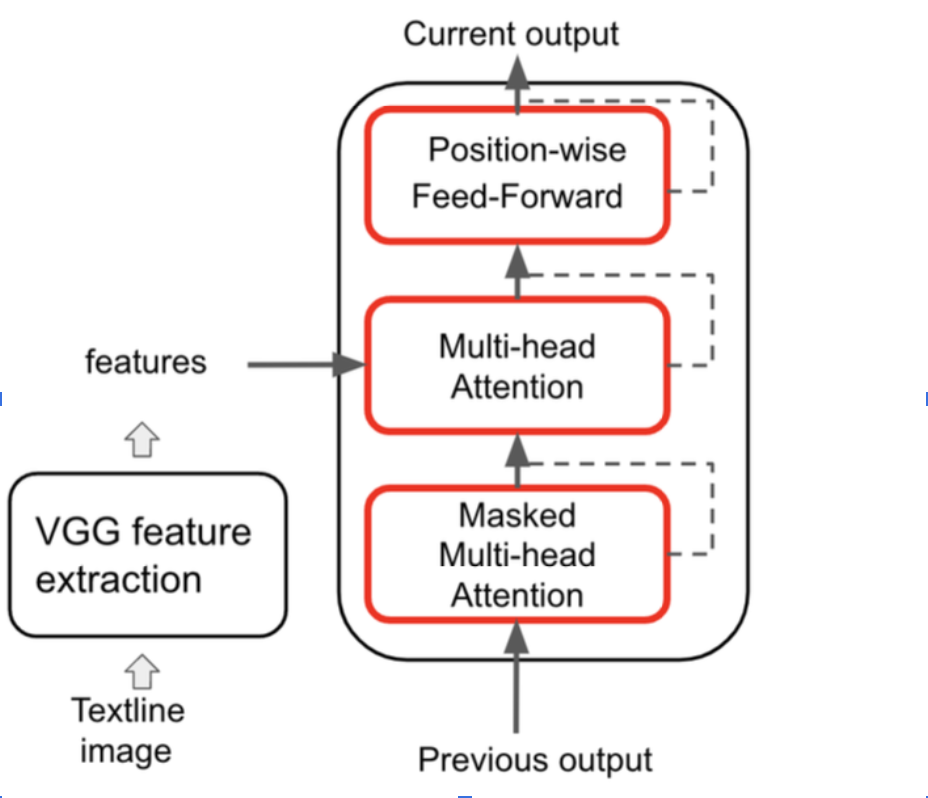}
\caption{Structure of the Transformer for recognizing Kindai documents.}\label{fig2}
\end{figure}

The baseline OCR is built on with the Transformer architecture, including a VGG feature extraction for extracting the visual features and a Transformer based decoder for generating the target text as shown in Figure~\ref{fig2}. We adopt the vanilla Transformer to generate the text with the guidance of the visual features and previous predictions.

Encoder: the encoder transforms the input textline image into a feature sequence representation. Based on the strong performance of convolutional neural architectures in feature extraction tasks, we employe VGG19 architecture to extract features \cite{bib7}. We remove the last two fully connected layers from the original VGG19 architecture and reuse the pretrained model from Imagenet. The output dimension from VGG19 is 512 channels. In addition to VGG19, we add a 1×1 convolution layer in the encoder part to adjust the image feature dimension to the size of the embedding dimension.

Decoder: the next character is predicted based on the input image and previously generated character. We employ the standard transformer model \cite{bib5}. Each transformer decoder layer module consists of three modules including: Masked Multi-head Attention, Multi-head Attention, Position-wise Feed-Forward as Figure~\ref{fig2}. The detailed implementation is described as follows:

Multi-head attention: employs multiple self-attention modules to have multiple representations of an input. The self-attention mechanism uses the query to obtain the value from key value pairs, based on their similarity.

Masked Multi-head attention: a mask matrix is used to enable the multi-head attention to restrict the attention region for each time step.

Position-wise Feed-Forward: has a linear layer, ReLU, and another linear layer, which processes each vector independently with identical weights.

\subsection{ Parallel Textline Images}\label{subsec3}

\begin{figure}[h]
\centering
\includegraphics[width=0.2\textwidth]{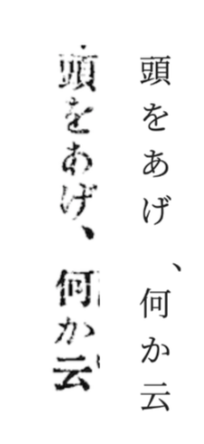}
\caption{Example of parallel textline images.}\label{fig3}
\end{figure}

The critical challenge in recognizing Kindai documents is the limited size of annotated datasets. The annotation process is labor-intensive and time-consuming to make transcriptions and bounding boxes for lines in documents. So, it is not scalable to create
a large dataset. As a result, the current public Kindai document dataset has only 4000 images, which is not efficient to train a large deep learning model for recognition. To enlarge the dataset, the general way is to employ a Japanese font to generate more samples. However, Kindai textlines have noise and damaged ink while font textlines are very clean. As a result, it is inefficient to train a Kindai OCR with font textline images. They are in different domains, so their representations will be separated. Thus, we cannot improve the Kindai OCR with font textline images if we train the Kindai OCR using the normal method. To overcome this problem, we generate parallel textline images to train a Kindai OCR. For each Kindai textline, we generate a font image with the same label. As the result, we have Parallel textline images that contain a pair of original Kindai and current Japanese fonts. Figure~\ref{fig3} shows an example of Parallel textline images, in which Kindai and font are on the left and right, respectively.

\subsection{ Adaptation Method}\label{subsec4}

\begin{figure}[h]
\centering
\includegraphics[width=0.7\textwidth]{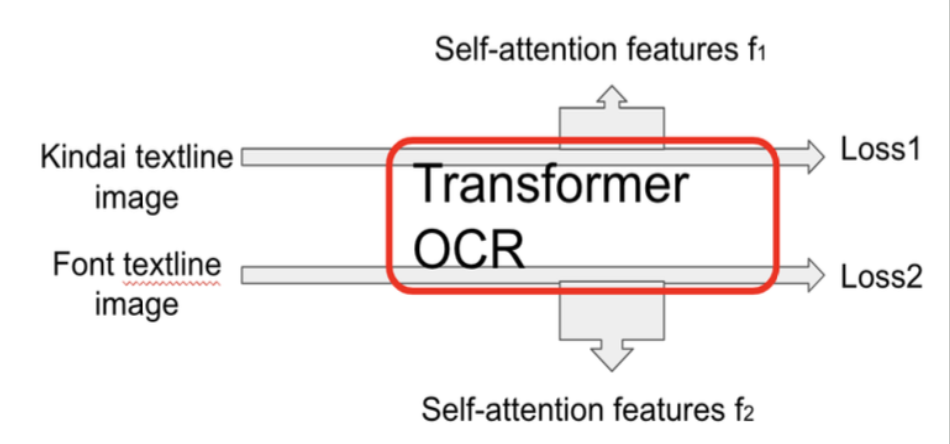}
\caption{Example of parallel textline images.}\label{fig4}
\end{figure}

We treat Kindai and font textline images as two different domains: Kindai and font. Our aim is to train the Kindai OCR with data from both domains and secure a good accuracy on the Kindai testing dataset. We need middle features extracted from the Kindai and font textline images to be similar. So while training we enforce Kindai OCR to extract similar features for both domain textline images. For parallel textline images  , we extracted two sequences of self-attention features  in the decoder as Figure~\ref{fig4}, where . Figure 4 shows the extracted self-attention features for parallel textline images. To enforce the Kindai OCR to extract similar sequential features for parallel textline images, we minimize the distance between two self-attention features. Hence it forces the Kindai OCR to extract domain-invariant features. In this research, we employ Euclidean distance and Maximum Mean Discrepancy to calculate the distance between two sequences of features.

\begin{figure}[h]
\centering
\includegraphics[width=0.6\textwidth]{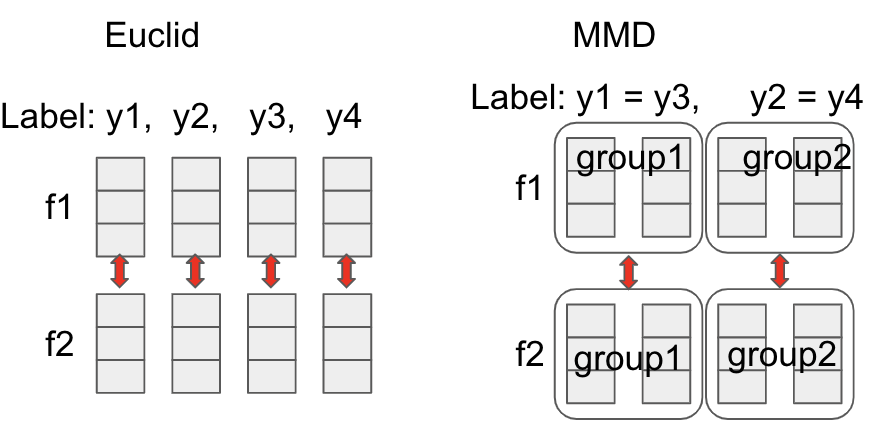}
\caption{Euclidean distance and MMD between 2 sequences of feature vectors.}\label{fig5}
\end{figure}

Euclidean distance: we calculate Euclidean distance between each pair of feature vector $(f_1^i, f_2^i)$. Then we take the average of those values through the sequence as Figure~\ref{fig5}.

Maximum Mean Discrepancy: is a statistical test used to determine where two distributions are the same. To calculate MMD between $(f_1, f_2)$, we first group feature vectors in $f_1$ and $f_2$ by character class $y_i$ . Each group contains the selt-attention features of the same character class and it represents the distribution of this character class. Then we calculate MMD for each group and take the average of value through the sequence as Figure~\ref{fig5}. We employ cross-entropy as the objective function to font textline image as $loss(x_1)$ and $loss(x_2)$ respectively. The distance between two self-attention features is added to enforce the system to extract similar features of parallel textline images.
\begin{equation}
 loss = loss(x_1) + loss(x_2) + * Distance((f_1, f_2) 
\end{equation}

Cross entropy loss ensures the Kindai OCR produces accurate predicted text lines by updating the weights of the entire system. In contrast, distance based loss ensures that features of Kindai and font domain are similar by updating weights of feature extractor and part of the decoder. During the inference time, we simply pass the Kindai textline image to Kindai OCR as usual.

\section{Experiments
}\label{sec4}

\subsection{Datasets}\label{subsec5}

\begin{figure}[h]
\centering
\includegraphics[width=0.6\textwidth]{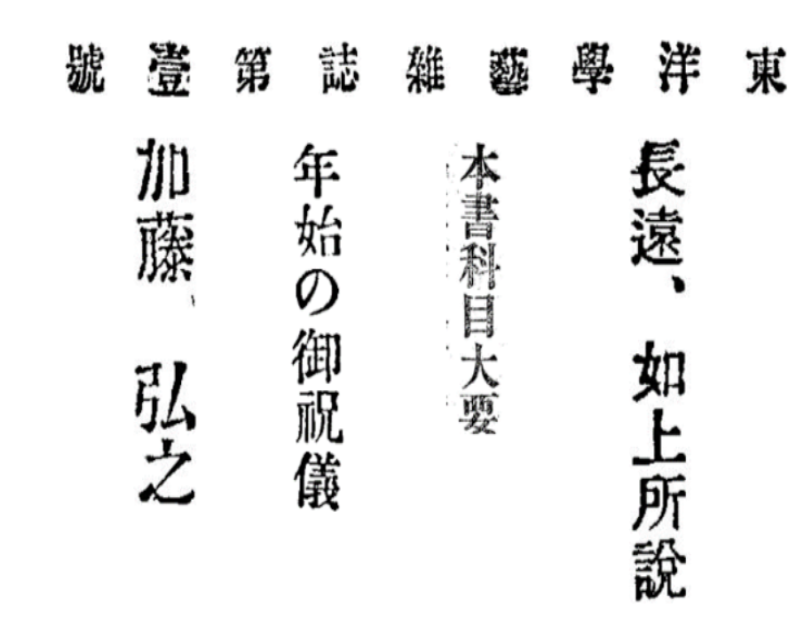}
\caption{Text line examples in Kindai documents.}\label{fig6}
\end{figure}

We employ annotated documents from the open dataset by the National Diet Library (NDL dataset) and the Shisou dataset in our previous work \cite{bib3} for this research. The NDL dataset has 3,997 images containing 10,421 and 92,835 horizontal and vertical textlines, respectively. The Shisou dataset has 922 images containing 38,770 textlines. The number of categories is 5,398, including many character categories that are not used in the current Japanese character system. Figure~\ref{fig6} shows some examples of vertical and horizontal text lines from Kindai documents. We employ the NDL dataset as training dataset and the Shisou dataset as evaluation dataset. For parallel text line images, we employ Noto Sans and Noto Serif to generate textline images. As a result, we increased the NDL dataset by 2 and 3 times, respectively.

For parallel texting images, we generate two datasets with different sizes. We employ Noto Sans font to generate font textline images, which we name dataset 1. The dataset has 103,256 parallel textline images. For the second dataset, we employ Noto Sans and Noto Serif to generate font textline images, which we name dataset 2. The number of parallel textline images in dataset 2 is twice as large as in dataset 1. We shuffle the Kindai and font textline images to train the Transformer OCR without adaptation, while we keep each pair of Kindai and font textline images to train the Transformer OCR with adaptation. Table~\ref{tab1} shows the number of images and parallel textline images of different datasets that we use in the experiments. For a fair comparison, we use the same number of images to train Transformer OCR with/without adaptation (note: dataset 2 contains duplicate original Kindai text line images for training Transformer OCR without adaptation).

\begin{table}[h]
\caption{The sizes of datasets to train Transformer OCR.}\label{tab1}%
\begin{tabular}{@{}lll@{}}
\toprule
Dataset & \# images & \# parallel textline images \\
\midrule
NDL dataset &  103,256 & N/A \\
P1 &206,512 &103,256 \\
P2 &413,024 &206,512 \\
\botrule
\end{tabular}
\end{table}

\subsection{Evaluation Metrics}\label{subsec6}
In order to measure the performance of our proposed method, we use the Character Error Rate (CER) metric which is generally employed for evaluating OCR systems. The detail of this metric is shown in the following equation:
\begin{equation}
 CER =  100* \frac{\sum_{i=1}^{n} ED (S_i, R_i)}{\sum_{i=1}^{n} |S_i|}\label{eq1}
\end{equation}
where $S_i$ is the $i-th$ string belonging the set of target strings (ground truth text) $S$. $R_i$ is the corresponding output string of the $S_i$ string. $|S_i|$ is the number of words in $S_i$. $ED(S_i, R_i)$ is the edit distance function which computes the Levenshtein distance between two strings $S_i$ and $R_i$.
\subsection{Implementation Details}\label{subsec7}

In the encoder part, we employ VGG19 with a 1×1 convolution layer to adjust the image feature dimension to 256 dimensions. In the decoder part, we use the standard transformer model. We set the embedded dimension and model dimension to 256, the number
of heads in the multi-head attention module to $H$= 12, and the number of transformer decoder layer to $N$= 12.

\subsection{Experimental Results}\label{subsec8}

The first experiment compares the performance of the Transformer OCR with the OCR system released by NDL (NDL OCR). NDL OCR employs a CRNN for textline recognition. Both systems are trained on the NDL dataset. Table~\ref{tab2} shows the CER on the Shisou dataset. Transformer OCR achieved 19.68\% of CER while the NDL OCR achieved 21.57\% on the Shisou dataset. This demonstrates the efficiency of Transformer architecture compared to CRNN architecture.

\begin{table}[h]
\caption{The CER of Transformer OCR and NDL OCR on the Shisou dataset.}\label{tab2}%
\begin{tabular}{@{}lll@{}}
\toprule
Method & Training dataset & CER on the testing set (\%) \\
\midrule
Transformer OCR  & NDL dataset &19.68 \\
NDL OCR & NDL dataset & 21.57 \\
\botrule
\end{tabular}
\end{table}

The second experiment evaluates the efficiency of parallel textline images and our proposed adaptation method. We train the Transformer OCR on different datasets with/without the proposed adaptation method. We employ single textline images to train the Transformer OCR and parallel texting images to train the Transformer OCR with the adaptation method using Euclidean distance and MMD. We employ the Transformer OCR trained on the NDL dataset as a baseline. The results in table~\ref{tab3} show that even if we enlarge the training dataset with font images, they do not help to improve the CER compared with the baseline. On the other hand, parallel textline images and the proposed adaptation methods improve 2.23\% and 3.94\% of CER with Euclidean distance and MMD, respectively.

\begin{table}[h]
\caption{The CER (\%) of Transformer OCR with/without adaptation method trained different datasets.}\label{tab3}%
\begin{tabular}{@{}llll@{}}
\toprule
Method & NDL dataset & P1 & P2 \\
\midrule
Transformer OCR & 19.68 & 19.55 & 19.59 \\
+ adaptation with Euclid &N/A &17.88 & 17.36 \\
+ adaptation with MMD & N/A & 15.93 & 15.65 \\

\botrule
\end{tabular}
\end{table}

\begin{figure}[h]
\centering
\includegraphics[width=0.8\textwidth]{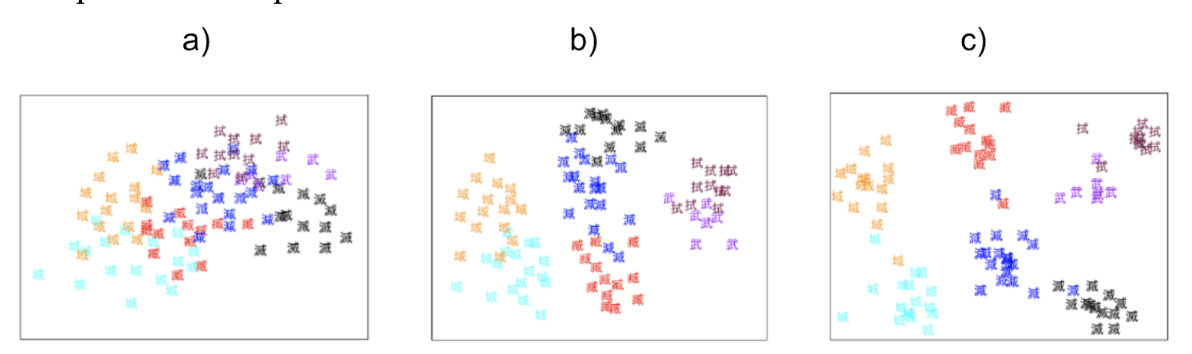}
\caption{Visualize the self-attention features on similar characters extracted by a) Transformer trained on P1 without adaptation. b) Transformer trained on P1 with adaptation + Euclidean distance. c) Transformer trained on P1 with adaptation + MMD.}\label{fig7}
\end{figure}

The model sizes of NDL OCR and Transformer OCR are 192 MB and 964MB respectively. It takes 13 hours and 15 hours to train the Transformer OCR without/with adaptation respectively on a GeForce GTX 1080 Ti. For inference, the inference speed of NDL OCR is 23ms per textline image while that of Transformer OCR is 160ms per textline image. The difference in inference speed is due to the much bigger model size of the Transformer OCR. We visualize the learned features between (a) Transformer OCR trained on P1 without adaptation and (b, c) Transformer OCR trained on P1 with adaptation with Euclidean distance and MMD, respectively. We select 7 similar characters and plot the final self-attention features of each character in the testing set. We employ t-SNE to reduce the feature vector from 256 dimensions to 2 dimensions as shown in Figure~\ref{fig7}. We observe that parallel textline images and the proposed adaptation method (b and c) provide a better representation of self-attention feature vectors than the original Transformer OCR (a). With MMD, we calculate the distance between distributions of characters rather than individual character samples as with Euclidean distance, so the distributions of characters in the same class are more compact and better separated from other classes. The OCR system learns invariant features between the Kindai and font domains, which helps the decoder produce better predictions.

\section{Conclusion}\label{sec5}

In this paper, we enlarged the small training set using parallel textline images and proposed a distance-based objective function to adapt between the Kindai and font domains. The self-attention feature-distance-based loss helps the OCR system produce the similar features between parallel textline images. The efficiency of the proposed adaptation and parallel textline images was demonstrated through experiments. Moreover, we visualize the feature vectors using Euclidean distance and MMD to understand how the system works. The Transformer OCR can learn invariant features across the Kindai and font domains, allow it to provide a better representation of characters in the middle feature space.

\bmhead{Acknowledgements}

The authors acknowledge Nguyen Tat Thanh University, Ho Chi Minh, Vietnam and CODH, Japan for supporting this study.

\bibliography{sn-bibliography_1}

\end{document}